\documentclass[smallcondensed]{svjour3}
\usepackage{mathptmx} % Times font
\usepackage{amssymb,amsmath}
\usepackage{stmaryrd}
\usepackage{verbatim}
\usepackage{graphicx}
\usepackage{url}
\usepackage{hyperref}
\usepackage{multicol}
\usepackage{multirow}
\usepackage{tikz}
\usepackage{makeidx}
\usepackage{pgfplots}
\usepackage{chngcntr}
\usepackage{marvosym}      % For letter symbol
\usepackage[round]{natbib} % For references options

\usetikzlibrary{shadows,arrows}
\hypersetup{
  colorlinks = true, % So the words are colored, not the boxes around them
  citecolor    = blue,
  linkcolor    = blue,
  urlcolor    = blue,
}
\setcitestyle{aysep={}}  % For avoiding commas in citations
\newcommand{\abscon}{\emph{Abscon}}
\newcommand{\argosmart}{ArgoSmArT-kNN}
\newcommand{\argocspknn}{ArgoCSP-kNN}
\newcommand{\azucar}{\textsf{Azucar}}
\newcommand{\bee}{\textsc{bee}}
\newcommand{\cphydra}{\textsc{CPhydra}}

\newcommand{\fzntwosmt}{\texttt{fzn2smt}}
\newcommand{\fzntwoxml}{\texttt{fzn2xml}}
\newcommand{\gecode}{\textsf{Gecode}}
\newcommand{\libsvm}{\texttt{LIBSVM}}
\newcommand{\mesat}{\emph{meSAT}}
\newcommand{\mistral}{\emph{Mistral}}
\newcommand{\mzncpx}{\texttt{mzn-g12cpx}}
\newcommand{\mznfd}{\texttt{mzn-g12fd}}

\newcommand{\mznlazy}{\texttt{mzn-g12lazy}}
\newcommand{\mznmip}{\texttt{mzn-g12mip}}

\newcommand{\mzntwofzn}{\texttt{mzn2fzn}}

\newcommand{\proteus}{\texttt{Proteus}}
\newcommand{\rapidminer}{\texttt{RapidMiner}}

\newcommand{\sugar}{\textsf{Sugar}}
\newcommand{\sunny}{\texttt{SUNNY}}

\newcommand{\xcsptwomzn}{\texttt{xcsp2mzn}}
\newcommand{\yices}{\textsf{Yices}}
\newcommand{\zthree}{\textsf{Z3}}

\newcommand{\Alldifferent}{\textit{all-different}}

\newcommand{\bestfixed}{\emph{best-fixed}}
\newcommand{\oracle}{\emph{oracle}}
\newcommand{\instance}{{instance}}
\newcommand{\knn}{\emph{k-NN}}

\newcommand{\minizinc}{MiniZinc}
\newcommand{\flatzinc}{FlatZinc}

\newcommand{\nSolvers}{15}
\newcommand{\preparationdata}{{preparation\_data}}
\newcommand{\solver}{{s}}
\newcommand{\solvers}{{solvers}}

\tikzstyle{file_}=[draw, fill=gray!10, text width=6.0em, text centered,
  minimum height=1.5em,drop shadow]
\tikzstyle{program_}=[draw, fill=gray!20, text width=6.0em, text centered,
  minimum height=1.5em,drop shadow]
\tikzstyle{file} = [file_, text width=8em, minimum width=10em,
  minimum height=3em, rounded corners, drop shadow]
\tikzstyle{filesmaller} = [file_, text width=6em, minimum width=7em,
  minimum height=5em, rounded corners, drop shadow]
\tikzstyle{program} = [program_, text width=6em, minimum width=10em,
  minimum height=3em, drop shadow]
\tikzstyle{linepart} = [draw, thick, color=black!50, -latex', dashed]
\tikzstyle{line} = [draw, thick, color=black!50, -latex']

\newcommand{\file}[2]{node (p#1) [file]{#2}}

\newcommand{\program}[2]{node (p#1) [program] {#2}}

\title{Short Portfolio Training for CSP Solving}
\author {Mirko Stojadinovi\'c \and Mladen Nikoli\' c \and Filip Mari\'c}

\institute{Mirko Stojadinovi\' c (\Letter) \at
              Faculty of Mathematics, University of Belgrade, Serbia\\
              \email{mirkos@matf.bg.ac.rs}\
           \and
           Mladen Nikoli\' c \at
           Faculty of Mathematics, University of Belgrade, Serbia\\
           \email{nikolic@matf.bg.ac.rs}\
           \and
           Filip Mari\' c \at
           Faculty of Mathematics, University of Belgrade, Serbia\\
           \email{filip@matf.bg.ac.rs}\
}

\date{Received: date / Accepted: date}

\journalname{Artificial Intelligence Review}

\begin{document}
\maketitle

\begin{abstract}
  Many different approaches for solving Constraint Satisfaction
  Problems (CSPs) and related Constraint Optimization Problems (COPs)
  exist. However, there is no single solver (nor approach) that
  performs well on all classes of problems and many portfolio
  approaches for selecting a suitable solver based on simple syntactic
  features of the input CSP instance have been developed. In this
  paper we first present a simple portfolio method for CSP based on
  k-nearest neighbors method. Then, we propose a new way of using
  portfolio systems --- training them shortly in the exploitation
  time, specifically for the set of instances to be solved and using
  them on that set. Thorough evaluation has been performed and has
  shown that the approach yields good results. We evaluated several
  machine learning techniques for our portfolio.  Due to its
  simplicity and efficiency, the selected k-nearest neighbors method
  is especially suited for our short training approach and it also
  yields the best results among the tested methods. We also confirm
  that our approach yields good results on SAT domain.
\end{abstract}


\setcounter{tocdepth}{2}

\pagestyle{empty}

% ------------------------------------------------------------------------------
\section{Introduction}
\label{sec:introduction}
% ------------------------------------------------------------------------------

\emph{Constraint satisfaction problems (CSPs)} and related
\emph{Constraint optimization problems (COPs)}
\citep{cp-book,cp-handbook} over finite domains are wide classes of
problems that include many problems relevant for real world
applications (e.g., scheduling, timetabling, sequencing, routing,
rostering, planning) \citep{cp-handbook}. Many different approaches for
solving CSPs exist (e.g., constraint propagation, backtracking
search algorithms, local search methods, constraint logic programming,
operation research methods, answer set programming, reduction to
SAT/SMT, lazy clause generation) \citep{cp-handbook} and there are
many state-of-the-art solvers that implement these approaches.

It is well-known that there is neither single solver nor single approach
suitable for all problems. When solving a CSP instance, one should
consider using several solvers and several different configurations
(setups) of each solver (if the solver is configurable). If a
multiprocessor machine is available, different solvers could be run in
parallel, until one of them solves the problem. However, in many cases
this is not feasible or desirable, so it is preferable to somehow
guess the solver that would give the best results for each given
instance and then run only that solver. \emph{Portfolio approaches}
that have been successfully used for SAT (e.g.,
\citep{satzilla,nikolic1,nikolic2,malitsky-knn,malitsky-cp11,isac,malitsky-cp12,malitsky-ijcai13}) but also for CSP
(e.g.,
\citep{cphydra,csp_portfolio2011,csp_portfolio2012,sunny,proteus})
assume that a number of different solvers are available and for each
input instance these approaches select a solver (and its
configuration) that should be run. This choice usually consists of
applying some machine learning technique and is most often based on
the knowledge gained during previous runs of the available solvers on
some training instances and on some syntactic characteristics of the
instance to be solved.

\emph{Finite linear CSP} \citep{sugar} is a special class of
constraint satisfaction problems that is often encountered in
applications. We consider only that class of problems, but also allow
global constraints \citep{gc-catalog}. In our previous work
\citep{mesat} we adapted the \argosmart{} portfolio \citep{nikolic2},
originally developed for SAT, to constraint satisfaction problems.
The main idea behind portfolio retains the simplicity of the original
approach, but the portfolio uses features of CSP instances and selects
CSP solver for the input CSP instance. In that paper, we considered
only CSP solvers based on reduction to SAT and we applied the
portfolio approach for selecting between different \emph{SAT
  encodings} that can be used for that reduction. In this paper, we
select between a very wide range of available solvers, including
solvers based on \emph{reduction to SAT} \citep{sat-handbook},
\emph{reduction to SMT} \citep{sat-handbook}, \emph{lazy clause
  generation solvers} \citep{lazy}, and \emph{constraint propagation}
solvers \citep{cp-handbook}. As different solving methods are usually
good at solving different types of problems, we want to exploit this
fact, thus increasing the potential efficiency of portfolio. We
compare the efficiency of our approach to the other state-of-the-art
CSP portfolios.

One of the problems with off-the-shelf application of portfolios is
that the portfolio may be trained on a set of instances with
properties significantly different than the set of instances to be
solved. In such cases the portfolio may perform poorly. Also,
retraining the portfolio for some specific set of instances may
require large amounts of time. We propose a new way of using
portfolios which relies on short portfolio training. Consider a
scenario in which a user wants to solve a specific, fixed set of
instances as fast as possible using solvers at his disposal. In our
approach, a short training run is performed on all of the given
instances in order to train a portfolio, and then, the portfolio is
used to solve those very same instances. Usually, when considering
portfolios (or machine learning methods and their applications in
general), one is concerned with their ability to generalize and would
not evaluate a portfolio on the instances it was trained on, as it
would give too optimistic estimate of its future performance. However,
if the goal is just to solve a specific set of instances, one need not
bother with portfolio's generalization ability and can just apply it
to the instances to be solved. To our knowledge, the portfolios have
not been used in this way before, mainly due to significant times
required by the standard portfolio training. Therefore, we provide a new way of
applying portfolios to solve practical problems.

We experimentally evaluate effectiveness of different machine learning
techniques when using short training in CSP portfolios. We evaluate
portfolios based on the k-nearest neighbors method,
support-vector machines, and linear regression.

Finally, we want to test if the main conclusions of our work transfer
to the SAT case.

Contributions of this work are the following.
\begin{itemize}
\item We present a thorough experimental evaluation of CSP solvers
  obtained by applying \nSolvers{} very diverse CSP solvers on the
  corpus containing more than 8,000 CSP instances.
\item We formulate \knn-based CSP portfolio and compare its
  effectiveness to other state-of-the-art portfolios -- our portfolio
  gives comparable results.
\item We test the effect of the training time on our \knn-based
  portfolio and show that it significantly improves performance over the
  best-fixed solver, even when extremely low solving timeouts are used
  (only several seconds per instance).
\item We propose a new way of using portfolio methods which relies on
  performing short training on the very same instances that should be
  solved.
\item We test effectiveness of using different machine learning
  techniques with short training and demonstrate that \knn-based
  portfolio gives the best results.
\item We show that main conclusions of our work on CSP transfer also
  to the case of SAT portfolios.
\end{itemize}

\paragraph{Overview of the paper.}
In Section \ref{sec:background} we give some basic definitions,
describe different solving methods and solvers, and present some of
the most well-known portfolio approaches. In Section
\ref{sec:solver_runs} we present experimental evaluation of different
CSP solvers on a large corpus of CSP instances. The results in Section
\ref{sec:solver_runs} are used as a basis for the next 3 sections. In
Section \ref{sec:portfolio} we describe our approach, compare it with
other most successful approaches and estimate the effect of the
training time on the portfolio effectiveness. The problem of obtaining
the best results on a fixed set of instances is addressed in Section
\ref{sec:short_training}. This section also provides comparison of
efficiency of different machine learning techniques using short
training and evaluation of the efficiency of our approach on SAT
problems. In Section \ref{sec:conclusions} we draw some final
conclusions and present ideas for further work.

% ------------------------------------------------------------------------------
\section{Background}
\label{sec:background}
% ------------------------------------------------------------------------------

In this section we introduce key notions used in the rest of the paper
and we analyze prior results in this area.

\subsection{Finite Linear CSP}

\begin{definition}
\label{def:def1}
\emph{Linear expressions} over the set of integer variables $V$ are
algebraic expressions of the form $\sum _{k=1}^{n} a_kx_k$
where all $x_k$ are variables from $V$ and all $a_k$ are integers.

A \emph{Finite Linear CSP in CNF} is a tuple $(V, L, U, B, S)$ where
\begin{enumerate}
\item $V$ is a finite set of integer variables,
\item functions $L: V \mapsto \mathbb{Z}$ and $U: V \mapsto
  \mathbb{Z}$ give lower and upper bound of integer variables and
  these bounds determine the domain $D(x)$ of each variable $x$,
\item $B$ is a set of Boolean variables,
\item $S$ is a finite set of clauses (over $V$ and $B$). Clauses are
  formed as disjunctions of literals where literals are the elements
  of the union of the sets $B$, $\{\neg p\,|\,p \in B\}$ and $\{e \leq
  c\,|\,e$ is linear expression over $V$, $c \in \mathbb{Z}\}$.
\end{enumerate}

A \emph{Solution} of Finite Linear CSP in CNF is an assignment of
Boolean values to Boolean variables and integer values to integer
variables satisfying their domains, such that all clauses from $S$ are
satisfied when variables are replaced by their values.
\end{definition}

\begin{example}
\label{exa:ex1}
A solution of Finite Linear CSP $V = \{x_1, x_2, x_3\}$, $L = \{x_1
\mapsto 1, x_2 \mapsto 1, x_3 \mapsto 2\}$, $U = \{x_1 \mapsto 2, x_2
\mapsto 4, x_3 \mapsto 3\}$, $B = \{p\}$, $C = \{ p\, \vee\, x_1 + x_3
\leq 4, \neg p\, \vee\, x_3 + (-1)\cdot x_1 \leq 0, x_1 \leq 1\,
\vee\, 2\cdot x_2 \leq 4\}$ is the assignment $\{p \mapsto \bot, x_1
\mapsto 1, x_2 \mapsto 3, x_3 \mapsto 2\}$.
\end{example}

In applications, the input syntax is usually modified so that it
allows non-contiguous domains, formulae with arbitrary Boolean
structure (not only CNF) and with literals formed by applying other
arithmetic operations (e.g., integer division, modulo) and other
arithmetic relations (e.g., $<$, $\geq$, $>$, $=$). All these formulae
alongside the clauses described in Definition \ref{def:def1} are
called \emph{intensional constraints}. Another usual modification of
the syntax is the usage of \emph{extensional constraints} (sometimes
called \emph{user-defined relations}) that are defined by a table of
allowed/disallowed assignments to the variables that they constrain. A
\emph{global constraint}\footnote{A catalogue of global constraints
  \citep{gc-catalog} is available online:
  \url{http://www.emn.fr/z-info/sdemasse/gccat}} is a constraint that
encapsulates a set of other constraints with two main purposes: to
increase expressiveness and improve efficiency by allowing specialized
algorithms for processing these constraints
\citep{regin_global}. Intensional, extensional, and global
constraints can be reduced to finite linear CSP in CNF form during
preprocessing, but usually more efficient procedures are obtained if
these are treated directly.

\begin{example}
\label{exa:ex2}
Constraint solver \sugar{} \citep{sugar} solves finite linear CSP by
reduction to SAT and it uses very simple input language. We give here
an example of finite linear CSP specification in this language.
\begin{verbatim}
(int x1 1 2) (int x2 1 4) (int x3 2 3)
(imp (>= (+ x1 (* 2 x3)) 3) (and (< x1 x2) (<= x3 (+ x1 x2))))
(alldifferent x1 x2 x3)
\end{verbatim}

The example uses intensional constraints and global constraint {\tt
  alldifferent}, stating that all its arguments have to take mutually
different values. The first row declares the domains of the variables
and the rest impose constraints on these variables. One of the
solutions to this problem is the assignment \verb|x1 = 1|,
\verb|x2 = 2|, \verb|x3 = 3|.
\end{example}

\subsection{Modeling languages}
Before solving, a constraint satisfaction problem must be somehow
specified. For this purpose, many modeling languages exist.  Two most
common modeling languages are \emph{\minizinc{}} and \emph{XCSP}.  We
also consider the \emph{\sugar{}} language used by several solvers in
our portfolio.

\minizinc{} \citep{minizinc} is a high-level constraint modeling
language. Before solving, most solvers compile \minizinc{}
specifications to a low-level target language \flatzinc{}. G12
\minizinc{} distribution contains many tools and solvers accepting
\minizinc{} and \flatzinc{} formats.

XCSP \citep{xcsp} is an XML-like low-level format used in several CSP
solving competitions.

Instances from XCSP format can be directly translated to \sugar{}
input format\footnote{Web page
  \url{http://bach.istc.kobe-u.ac.jp/sugar/current/docs/syntax.html}
  contains full syntax of \sugar{} input language.}, that has much
simpler syntax (Example \ref{exa:ex2}).

\subsection{Solving methods for CSPs}

\paragraph{Propagation-based systems.}
The process of \emph{Constraint propagation} \citep{cp-handbook}
reduces a CSP to an equivalent problem, simpler to solve. The
reduction is most often done by removing values from the domains of
the variables that cannot be the part of any solution to the
problem. Many solvers that are based on constraint propagation
techniques were developed. These solvers use search heuristics and
algorithms for performing constraint propagation, thus achieving
different types of consistencies \citep{cp-book,cp-handbook}. Solvers
\mistral{} \citep{mistral} and \abscon{} \citep{abscon} that
participated in CSP competitions belong to this type of solvers.

\paragraph{Reduction to SAT.}
\emph{Propositional satisfiability problem (SAT)} \citep{sat-handbook}
is the problem of deciding if there is a truth assignment under which
a given propositional formula (in conjunctive normal form) evaluates
to true. It is a canonical NP-complete problem \citep{cook} and it
holds a central position in the field of computational complexity.
When using reduction to SAT, CSP instances are encoded as SAT
instances and modern efficient satisfiability solvers are used for
finding solutions that are then converted back to the solutions of the
original CSPs.

A fundamental design choice when encoding finite domain constraints
into SAT concerns the representation of integer variables. Several
different encoding schemes \citep{compact-order,mesat} have been
proposed and successfully used in various applications. There are
several tools that reduce CSPs to SAT using one or more of standard
encodings (e.g., direct encoding, support encoding, order encoding,
log encoding).

\sugar{} is a constraint solver that solves finite linear CSPs by
translating them into SAT using the order encoding method \citep{order}
and then solving SAT instances by several supported SAT solvers.

\azucar{} \citep{azucar} is a successor of \sugar{} that uses the
compact-order encoding \citep{compact-order} for translating finite
linear CSP into SAT. It is tuned for solving specific large domain
sized CSP instances. Log encoding is a special case of compact-order
encoding so \azucar{} can also use this encoding when reducing to
SAT.

\mesat{} \citep{mesat} (Multiple Encodings of CSP to SAT) is a system
using different encodings and their combinations. The supported
encodings are: direct \citep{direct}, support \citep{support},
direct-support \citep{mesat}, order \citep{order} and direct-order
\citep{mesat}.

\paragraph{Reduction to SMT.}
\emph{Satisfiability modulo theories (SMT)} \citep{sat-handbook} is a
research field concerned with the satisfiability of formulae with
respect to some decidable first-order background theory (or
combination of them). Some of these theories are \emph{Linear Integer
  Arithmetic}, \emph{Integer Difference Logic}, \emph{Linear Real
  Arithmetic}, etc. There are several systems that solve CSPs
by reduction to SMT (e.g. \fzntwosmt{} \citep{fzn2smt}).

\paragraph{Lazy clause generation.}
In the \emph{lazy clause generation} approach \citep{lazy}, finite
domain constraint propagation engine is combined with a SAT solver:
propagators are mapped into clauses and passed to SAT solver, which
uses unit propagation and then returns obtained information back to
the engine. Contrary to the eager approach, clauses are not generated
a priori but are constructed and given to the SAT solver during the
solving phase. The lazy propagation approach can be viewed as a
special form of Satisfiability Modulo Theories solver, where each
propagator is considered as a separate theory, and theory propagation
is used to learn clauses. Solvers \mzncpx{} and \mznlazy{} (included
in the \minizinc{} G12 distribution) implement lazy clause generation.

\subsection{Machine learning techniques used for CSP portfolio}

Machine learning methods are often used to predict value of some outcome
variable based on the values of some other
related variables, called {\em features}. There are various methods
that address this task. Usually, some form of statistical model that
expresses the dependence of the outcome variable on feature variables
is assumed. The coefficients of the model are determined using a set
of instances for which both the values of outcome and feature variables
are known. We describe some frequently used methods that are used
in this paper.

\emph{Linear regression} is one of the most often used machine
learning methods. The variable to be predicted is modeled by a linear
combination of feature variables \citep{bishop_pattern}. The
coefficients of the model can be determined by the well known least
squares method in order to minimize sum of square differences between
the predicted and actual outcomes on training data.

In \emph{k-nearest neighbors method}, it is assumed that the outcome
variable is modeled based on its values on $k$ instances from the
training set that are closest to the input instance, with respect to
some distance measure defined over feature vectors. Number $k$ is
selected empirically.  If the outcome variable is continuous, its
value is taken to be a linear combination of its values on the
neighbor instances. For the coefficients of the linear combination the
simplest choice is $1/k$, but they can also depend on the distances
involved. If the outcome variable is discrete, its value is decided by
voting of neighbor instances.

In its basic form, \emph{support vector machine (SVM) regression} models
the outcome variable by a linear combination of
feature variables with additional requirements. Firstly, the
predictions need to deviate from actual values of the outcome variable
by at most $\varepsilon$ (which is a metaparameter) and secondly, the
model should be as flat as possible.  Flatness is ensured by
minimizing the norm of the coefficient vector. Modifications are made
to tackle the cases in which there is no function which fits the
deviation constraints and to allow for nonlinear regression functions.

Training algorithms usually have some metaparameters which influence
the model that the algorithm yields for specific data set. Their
selection is a nontrivial process and the best choice of the values
for metaparameters is usually made by evaluating the models that
different values of metaparameters yield.

In order to check if models produced by machine learning methods
generalize well, they should be evaluated on a data set different than
the training set. The evaluation is usually performed in one of two
ways. First one is to split the available data to training and test
set, perform training and produce a model on training set, and then
measure how well it predicts the outcome variable in the test set. In
this scenario, the question arises how to select the test
set. Different splits to training and test set may result in different
evaluation results.  To avoid that, second evaluation scenario ---
k-fold cross-validation --- is used.  The available data is split to
$k$ parts (called folds). For each part, the model is trained on $k-1$
other parts and evaluated on that part. At the end, evaluation results
are aggregated. Cross-validation produces more reliable estimate of
the generalization capability of the model, but is clearly more time
consuming.

\subsection{Solver selection for SAT and CSP}

\paragraph{SAT portfolios.}
The instance-based solver selection problem has been widely studied in
the SAT community. Based on the characteristics of the input instance,
either some parameters of a single solver are tuned, or one of several
available solvers is selected to be applied on that instance
(so called solver portfolio). The most successful results are based on
machine-learning techniques (e.g., SATZilla \citep{satzilla},
\argosmart{} \citep{nikolic1,nikolic2}, ISAC \citep{isac}, Non-Model-Based
Algorithm Portfolios for SAT \citep{malitsky-knn}). Each SAT instance
is characterized by a set of its features (most of them are purely
syntactic and extracted from the CNF representation). Usually, a
training corpus is solved by different SAT solvers (or a single solver
configured by different parameters) and for each instance, its solving
times for each solver are assigned to its feature vector.
For each solver, a predictive model is learned, which
describes the dependence between feature vectors and solving times.
When a new instance is to be solved, the most suitable solver
is chosen, based on solving times predicted by models of
each solver for feature vector of the input instance.

\paragraph{CSP portfolios.}
Algorithm portfolios have recently been applied to constraint
satisfaction problem solving. \cphydra{} \citep{cphydra} is an
algorithm portfolio for CSP that uses k-nearest neighbors algorithm to
determine one or more solvers to be run on an unseen problem
instance. The superiority of the portfolio over each of its
constituent solvers is demonstrated using challenging benchmark
problem instances from CSP Solver Competition. Another approach,
described by Kiziltan et al.~\citep{csp_portfolio2011}, uses run-time
classifiers (categories are: ``short'', ``medium'' and ``long'') to
minimize the average solving time of each instance. This portfolio
uses features of \cphydra{} and SATZilla and the combination of the
two. Work by Amadini et al.~\citep{csp_portfolio2012} compares
efficiency of different portfolio approaches based on SAT portfolio
techniques and machine learning algorithms.

A recent system \sunny{} \citep{sunny} outperforms \cphydra{} and some
portfolios originally developed for SAT, but adapted to CSP solving
(e.g. SATZilla). For some new instance to be solved, \sunny{} uses
\knn{} algorithm to select one or more solvers that should be
applied. In practice, the number of selected solvers is very rarely
greater than two. Selected solvers share available time (each selected
solver is assigned certain amount of time from the total time
available, based on the efficiency of each solver on \verb|k| nearest
instances). \sunny{} uses 155 features, gathered from instances in the
\minizinc{} input format, using tool \emph{mzn2feat} developed by the
same authors. From these features, 11 are dynamic, i.e. obtained by
running \gecode{} solver for 2 seconds. The original experimental
evaluation \citep{sunny} included instances from CSP Solver
Competition and from \minizinc{} distribution, and 11 solvers from
\minizinc{} Challenge.

\proteus{} \citep{proteus} is a hierarchical portfolio that outperforms
many other machine learning approaches adapted for CSP solving. When a
new instance is to be solved, a solver is selected by making decisions
on two or three levels. On the first level, only the solving approach
is chosen -- either (not SAT-based) CSP solving or reduction to
SAT. If CSP solving is chosen, the CSP solver that should be applied
is selected on the second level. If reduction to SAT is chosen, only
the encoding method is chosen on the second level, and the solver is
selected separately, on the third level. As stated by the authors, the
advantage of this approach is that most suitable technique can be used
for each decision level, and this can lead to improved performance. On
the other hand, usage of different techniques at each level means the
system is of a much greater complexity than some other
approaches. \proteus{} uses 36 CSP features, the same as \cphydra{},
and among these features some dynamic, obtained by running \mistral{}
solver for 2 seconds. For each of 3 SAT encodings, \proteus{} uses 54
SAT features. Instances from CSP Solver Competition were used and
\proteus{} was tested with 4 CSP solvers, 3 SAT encodings and 6 SAT
solvers.

In our previous work \citep{mesat}, we described portfolio for
selecting between different SAT encodings used by system \mesat. The
selection process is based on \knn{} algorithm and the constructed
portfolio outperformed all the constituent encodings (direct-support,
order, direct-order) used in the experiments. One of the assumptions
of that paper was that the most efficient encoding in solving easier
instances of some family of instances (coming from a single problem)
will also be the most efficient in solving harder instances of the
same family. On the instances used in the experiments, this assumption
was demonstrated to be right and the experiments showed that portfolio
\emph{trained only on easy instances} (solvable by all used encodings
in some small time) can outperform any constituent encoding.

This paper is a continuation of our previous work: the same portfolio
is used, but now with different CSP solvers and not only different
encodings. We provide the pseudocode of the portfolio and describe it
more formally. The ways the portfolios are used in the original paper
and in this paper significantly differ -- in this paper there is no
assumption that instances are divided in families, the training is not
performed only on easy instances, the aims of experiments are
different, etc.

% ------------------------------------------------------------------------------
\section{Experimental Setup}
\label{sec:solver_runs}
%------------------------------------------------------------------------------

In this section, we describe our experimental setup that will be used
throughout the paper. We use a rich set of available CSP solvers and a
very wide corpus of available CSP benchmarks. Aside of describing
them, here we also present the evaluation of these solvers on these
benchmarks.

\paragraph{Solvers.}
For reduction to SAT, we used direct-support, order and direct-order
encodings implemented in system \mesat{} 1.1 \citep{mesat}, order
encoding implemented in system \sugar{} 2.1.3 \citep{sugar}, and log
and compact-order encoding implemented in system \azucar{} 0.2.4
\citep{azucar}. SAT solver Minisat 2.2 \citep{minisat} was used in all
cases when reduction to SAT was performed. We extended \mesat{} to
enable reduction of CSP instances to SMT-LIB language\footnote{The
  source code of our implementation and the instances used in
  experiments (but without third-party solvers, due to specific
  licensing) are available online from \mesat{} web page:
  \url{http://jason.matf.bg.ac.rs/~mirkos/Mesat.html}}. The
translation of most constraints is straightforward, and only global
constraints are decomposed to more simpler constraints. SMT solvers
\yices{} 2.2.0 \citep{yices} and \zthree{} 4.2 \citep{z3} were used
for solving generated SMT-LIB instances. We used two solvers
implementing constraint propagation techniques: \abscon{} 112V4
\citep{abscon} and \mistral{} 1.545 \citep{mistral}. We also used
solvers from G12 \minizinc{} 1.6 \citep{minizinc} distribution:
\mznlazy{} and \mzncpx{} implementing lazy clause generation, and
\mznfd{} and \mznmip{}.  Solver \gecode{} 4.2.1 \citep{gecode} was
also used.

\paragraph{Instances.}
We used two publicly-available corpora of CSP instances: (i) CPAI09
corpus containing all instances used in Fourth International CSP
Solver
Competition\footnote{\url{http://www.cril.univ-artois.fr/CPAI09}},
(ii) instances from \minizinc{} corpus available on \minizinc{}
page\footnote{\url{http://www.minizinc.org}}. We used three formats of
input files: \minizinc{} language \citep{minizinc}, XCSP format
\citep{xcsp} and \sugar{} \citep{sugar} input format.

\begin{figure}
  \centering
  \begin{tikzpicture}[scale=0.8,transform shape]

  % Left diagram
  \path \file{1}{XCSP format};
  \path (p1.south)+(-1.5,-1.1) \program{2}{\xcsptwomzn{}};
  \path (p1.south)+(1.5,-1.1) \program{3}{\sugar{} converter};
  \path (p2.south)+(0,-1) \file{4}{\minizinc{} format};
  \path (p3.south)+(0,-1) \file{5}{\sugar{} format};
  % Right diagram
  \path (p1.east)+(4,3) \file{6}{\minizinc{} format};
  \path (p6.south)+(0,-0.9) \program{7}{\mzntwofzn{}};
  \path (p7.south)+(0,-0.9) \file{8}{\flatzinc{} format};
  \path (p8.south)+(0,-0.9) \program{9}{\fzntwoxml{}};
  \path (p9.south)+(0,-0.9) \file{10}{XCSP format};
  \path (p10.south)+(0,-0.9) \program{11}{normalization};
  \path (p11.south)+(0,-0.9) \file{12}{XCSP format};
  \path (p11.east)+(2.5,0) \program{13}{\sugar{} converter};
  \path (p13.south)+(0,-0.9) \file{14}{\sugar{} format};

  % Left lines
  \path [line] (p1.south) -- +(0.0,-0.2) -- +(-1.5,-0.2) -- node [above, midway] {} (p2);
  \path [line] (p1.south) -- +(0.0,-0.2) -- +(+1.5,-0.2) -- node [above, midway] {} (p3);
  \path [line] (p2.south) -- node [above, midway] {} (p4);
  \path [line] (p3.south) -- node [above, midway] {} (p5);
  % Right lines
  \path [line] (p6.south) -- node [above] {} (p7);
  \path [line] (p7.south) -- node [above] {} (p8);
  \path [line] (p8.south) -- node [above] {} (p9);
  \path [line] (p9.south) -- node [above] {} (p10);
  \path [line] (p10.south) -- node [above] {} (p11);
  \path [line] (p11.south) -- node [above] {} (p12);
  \path [line] (p12.east) -- +(0.3,0) -- +(0.3,1.35) -- node [above, midway] {} (p13);
  \path [line] (p13.south) -- node [above] {} (p14);

  \end{tikzpicture}
  \caption{Conversions between different input formats.}
  \label{fig:conversions}
\end{figure}
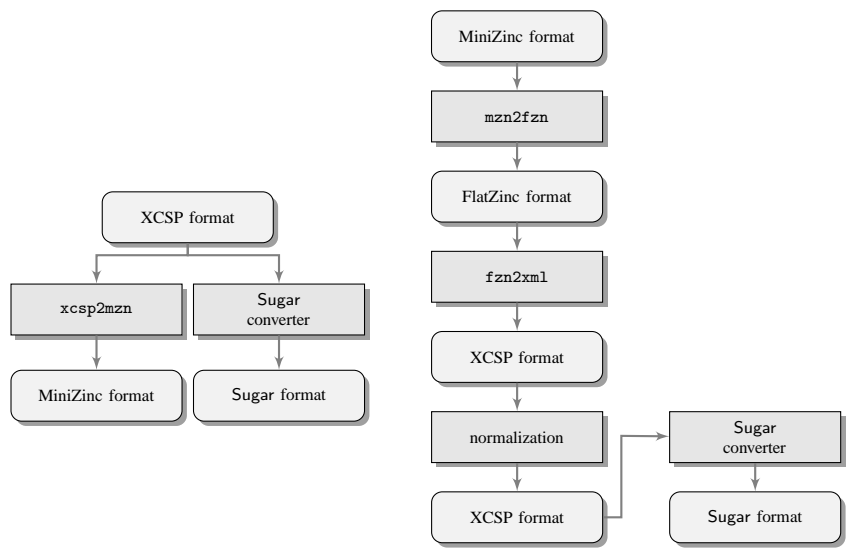

Instances are converted between different formats using of-the-shelf
tools and the conversions are presented in Figure
\ref{fig:conversions}. Instances from the first corpus were
automatically converted from the original input language to
\minizinc{} by the converter \xcsptwomzn{} available on \minizinc{}
page and to \sugar{} input format by the converter included in the
\sugar{} distribution. Instances from the second corpus use model
files and data files from \minizinc{} distribution corpora. These
instances were translated to FlatZinc format by \mzntwofzn{}, then
from this format to XCSP format by \fzntwoxml{} (both converters are
available in \minizinc{} distribution) and then normalized by the tool
provided by Amadini et al. \citep{csp_portfolio2012}. From XCSP format
these instances were also converted to \sugar{} input format.

We excluded instances that could not be converted from any of two
formats to any other format and instances for which there was a solver
giving wrong satisfiability answer (surprisingly, there were 541
instances where at least one of the solvers gave wrong answer, and 8
out of 15 solvers gave wrong answers on some instances). The final
corpus consisted of 8436 instances.

\paragraph{Experimental environment.}
All tests were performed on a multiprocessor machine with AMD
Opteron(tm) CPU 6168 on 1.9Ghz with 2GB of RAM per CPU, running
Linux. Solving timeout was 600 seconds for each instance (for total
time including selecting the solver where needed, encoding where
needed, and solving).

\paragraph{Solver comparison.}
We exhaustively ran all solvers on all instances. The results are
shown in Table \ref{tbl:solvers}. The number of solved instances and
total solving time (in days) are shown (for each unsolved instance,
the timeout value of 600 seconds is assumed).  Mean
time per instance is directly computable from the total time and the
number of instances. We give results for two additional methods: (i)
\oracle{} (or virtual best) method that would select the best solver
for each instance (this method is not feasible in practice since
for each instance it must always guess the best solver)
and (ii) the \bestfixed{} method -- one solver that gives
the best overall performance. The results indicate that the difference
between the \bestfixed{} and the \oracle{} is 1415 instances, so, it makes
sense to apply a portfolio approach.

\begin{table}
\begin{center}
\begin{tabular}{|l|l|c|c|}\hline

Solving method & Solver & \# (out of 8436) & Time (days) \\\hline
\multirow{2}{*}{Propagation based}
& \abscon{}                 & 4402 & 30.2 \\
& \mistral{}                & 6216 & 16.9 \\\hline
\multirow{7}{*}{Reduction to SAT}
& \mesat{} (direct-support) & 3128 & 38.6 \\
& \mesat{} (order)          & 3985 & 32.6 \\
& \mesat{} (direct-order)   & 3984 & 32.6 \\
& \sugar{} (order)          & 5548 & 22.7 \\
& \azucar{} (compact-order) & 4402 & 31.1 \\
& \azucar{} (log)           & 3804 & 30.7 \\\hline
\multirow{2}{*}{Reduction to SMT}
& \zthree{}                 & 4938 & 26.7 \\
& \yices{}                  & 4978 & 26.0 \\\hline
\multirow{2}{*}{Lazy clause generation}
& \mzncpx{}                 & 2883 & 39.4 \\
& \mznlazy{}                & 2865 & 41.3 \\\hline
\multirow{3}{*}{Other}
& \mznmip{}                 & 1544 & 49.0 \\
& \mznfd{}                  & 2716 & 41.7 \\
& \gecode{}                 & 2627 & 41.4 \\\hline
\multirow{2}{*}{}
& \bestfixed{}              & 6216 & 16.9 \\
& \oracle{}                 & 7631 & 7.0 \\\hline
\end{tabular}
\end{center}
\addtolength{\belowcaptionskip}{3mm}
\addtolength{\abovecaptionskip}{-2mm}
\caption{Results of experimental evaluation of CSP solvers;
  the encoding used for reduction to SAT is given in parentheses; \#
  denotes the number of solved instances; \bestfixed{} --  single
  solver achieving the best results (\mistral{} in this case),
  \oracle{} -- the best solver is chosen for each instance.}
\label{tbl:solvers}
\end{table}

% ------------------------------------------------------------------------------
\section{CSP Portfolio}
\label{sec:portfolio}
%------------------------------------------------------------------------------

In this section we present a simple portfolio for CSP. The basic
principle was taken from SAT portfolio \argosmart{} \citep{nikolic2},
but the new portfolio was constructed based on features (described in
Section \ref{sec:features}) and solvers suited for CSPs. In
Section \ref{sec:argocspknn} we describe the portfolio, in Section
\ref{sec:other_approaches} we compare efficiency of \argocspknn{} to
different, already existing, portfolios. Finally, in Section
\ref{sec:autocv} we briefly evaluate the effect of instance solving time
on portfolio effectiveness.

\subsection{Portfolio features}
\label{sec:features}

Unlike some other approaches (e.g., \citep{csp_portfolio2011}) that
use features of the generated SAT instances (in case of reduction to
SAT), we use only features extracted from the original CSP
instance. We considered 70 different features\footnote{The already
  mentioned \mesat{} web page contains a detailed description of all
  70 features.} divided in several groups: features related to the
sizes of the domains of integer variables for all variables in the
instance (e.g., average domain size), and for the variables included
in each different type of constraint, features related to the number
of all variables and variables with non-contiguous domains, features
related to the number and the percentage of the constraints of
different types --- intensional (e.g., percentage of intensional
constraints among all the constraints), extensional, global (e.g.,
average arity of global constraints), as well as for each specific
type of constraint (e.g., number of arithmetic constraints, number of
multiplications, sum of domains of variables involved in
multiplications, number of \Alldifferent{} constraints), etc.

In most of the experiments -- except when we explicitly state that
we did differently -- we used all 70 features. The time used for
the feature extraction is small (about 0.05 seconds in average on all
instances).

\subsection{\argocspknn{} portfolio}
\label{sec:argocspknn}

Before it can be applied, \argocspknn{} portfolio must be trained.
{\em Portfolio training} consists of solving the training corpus ({\em the preparation})
with all available solvers and {\em predictive model training}. Once trained,
portfolio can be used many times. That phase we call {\em exploitation} phase.

\paragraph{Preparation.}
First, the features of all training instances and performance of all
solvers on these instances need to be gathered. For solving each
training instance, each solver is given certain amount of time, called
the \emph{solving timeout}. Solver performance on an instance is
expressed by a \emph{PAR10} score \citep{par10} --- the solving time if
the instance is solved within the given timeout, or the timeout value
multiplied by 10, otherwise. Results of preparation phase are
expressed by two tables. For each instance, the first table contains
its features. The second table contains the PAR10 score for the given
solving timeout for each solver and each instance.

\paragraph{Predictive model training.}
The purpose of this phase is to select (based on the results of the
preparation phase) the optimal values for $k$ (the number of
neighbors) and $d$ (the distance measure) that are later going to be
used in the portfolio exploitation. This is done by using the
\emph{cross-validation} technique. Different values of $k$ and $d$ are
tried. For each fixed combination of these two metaparameters, a
5-fold cross-validation on prepared data is performed. At each turn,
for each instance of one fold, its $k$ nearest neighbors (with respect
to the distance measure $d$) are found among instances of four other
folds.  The solver having the smallest sum of PAR10 scores on these
neighboring instances is considered the best and selected. The
combination of $k$ and $d$ giving the best results on all training
instances (the smallest sum of PAR10 scores of selected solvers for
all training instances) is the one that is going to be used by the
portfolio.

\paragraph{The portfolio and its exploitation.}
The solver selection algorithm (based on \citep{nikolic2}) is shown in
Figure \ref{fig:argocspknn}. The values for the input parameter
\verb|preparation_data| are collected during preparation, while the values
for input parameters \verb|k| and \verb|d| are chosen in the process
of predictive model training. In the exploitation phase, first the
features of the input instance are computed. Then, the nearest
neighbors of the input instance in the whole training corpus are found
and the best solver on these instances is used to solve the input
instance. Function {\tt par10} returns the sum of PAR10 scores of some
specific solver on neighboring instances.

\begin{figure}
{\tt
\noindent
\instance{} - instance to be solved\\
\solvers{} - set of solvers\\
\preparationdata{} - features and PAR10 scores of solvers on training instances\\
k - number of neighbor instances to be considered\\
d - distance measure\\
\\
{\bf ArgoCSP-kNN} (\instance{}, \solvers{}, \preparationdata{}, k, d)\obeylines
\quad features = CSPFeatures (\instance);\obeylines
\quad neighbors = nearestNeighbors (\preparationdata{}, k, d, features);\obeylines
\quad best\_score = $\infty$;\obeylines
\quad {\bf for} \solver{} {\bf in} \solvers{} \obeylines
\quad \quad {\bf if} (par10 (\solver{}, neighbors) $<$ best\_score) \obeylines
\quad \quad \quad best\_score = par10 (\solver{}, neighbors);\obeylines
\quad \quad \quad best\_solver = \solver{};\obeylines
\quad {\bf return} best\_solver;\obeylines
}
\caption{\argocspknn{} solver selection procedure{}.}
\label{fig:argocspknn}
\end{figure}

% ------------------------------------------------------------------------------
\subsection{Comparison with other CSP portfolios}
\label{sec:other_approaches}
%------------------------------------------------------------------------------

In this subsection we compare \argocspknn{} with other CSP
portfolios. Since the most important part of our work is the
methodology based on very short timeouts (presented in
Section \ref{sec:short_training}), and not the portfolio itself, our
aim is not to outperform the existing approaches by using
\argocspknn{}, but to show that it is comparable with them. The
comparison is performed with \sunny{} and \proteus{}, two portfolios
that outperformed many other developed approaches.

\paragraph{Comparison with \sunny{}.}
\argocspknn{} and \sunny{} use different sets of solvers, features and
instances, so a direct comparison of the two approaches was not
possible without adapting one of them to use the same features,
solvers and instances as the other. We applied the \sunny{} approach
and implemented a \sunny{}-like portfolio that uses the same solvers
and features\footnote{We also experimented with {\tt mzn2feat} tool
  for collecting features provided by the authors of \sunny{}, but it
  was very slow on the instances used in our experiments and could not
  collect features of all instances even after several days.} as
\argocspknn{} and evaluated it on our rich experimental corpus. Apart
from different solver scheduling, another difference between \sunny{}
and \argocspknn{} is that \sunny{} uses some fixed number of neighbors
$k$ and some fixed distance measure $d$, and our approach determines
the optimal values of these metaparameters in the model training
phase. The number of neighbors ($k=16$) that produced the best results
in the experiments in the original paper was used. The Euclidean
distance measure was used as suggested by the authors. We used 5-fold
cross-validation evaluation scheme.  The results of comparison between
our method and \sunny{} are given in Table \ref{tbl:sunny_comparison}.

 \begin{table}
 \begin{center}
 \begin{tabular}{|l|c|}\hline
         & Solved instances (out of 8436) \\\hline
 \bestfixed{}  & 6216 \\\hline
 \sunny{}      & 7493 \\\hline
 \argocspknn{} & 7511 \\\hline
 \oracle{}     & 7631 \\\hline
 \end{tabular}
 \end{center}
 %\addtolength{\belowcaptionskip}{-8mm}
 \addtolength{\abovecaptionskip}{-2mm}
 \caption{Results of experimental comparison
   of \argocspknn{} and \sunny{}.}
 \label{tbl:sunny_comparison}
 \end{table}

 Results show that on the used corpus our approach is comparable with
 \sunny{}.  Slight improvement may be due to flexibility of our
 approach with respect to selection of $k$ and $d$, but since the
 improvement is rather small, we speculate that the situation might
 even be opposite on different instances, and that no significant
 difference in performance exists.

\paragraph{Comparison with \proteus{}.}
As features of used instances and the times for all solvers on all
instances used in \proteus{} paper \citep{proteus} had been made
available online by the authors, we decided to apply the \argocspknn{}
approach to that data. All instances in the \proteus{} corpus satisfy
two criteria: they are not trivially solved during 2 seconds of
feature computation and they are solved by at least one of the used
solvers within the time limit of 1 hour. The set of solvers included 4
CSP solvers and 6 SAT solvers, and three different encodings (direct,
support and order) were used when reducing to SAT. For \argocspknn, we
did not use features of generated SAT instances but only 36 features
obtained from the input CSP instances. The results of comparison
between \argocspknn{} and \proteus{} are given in Table
\ref{tbl:proteus_comparison}. The number of solved instances of
\proteus{} is taken from the paper introducing this portfolio. As in
the case of \sunny{} the difference is very small, so we assume that
two systems perform roughly the same.

 \begin{table}
 \begin{center}
 \begin{tabular}{|l|c|}\hline
         & Solved instances (out of 1493) \\\hline
 \bestfixed{} & 984 \\\hline
 \argocspknn{}  & 1413 \\\hline
 \proteus{}  & 1424 \\\hline
 \oracle{} & 1493\\\hline
 \end{tabular}
 \end{center}
 %\addtolength{\belowcaptionskip}{-8mm}
 \addtolength{\abovecaptionskip}{-2mm}
 \caption{Results of experimental comparison
   of \argocspknn{} and \proteus{}.}
 \label{tbl:proteus_comparison}
 \end{table}

\bigskip

Note that \argocspknn{} solved respectively 98.4\% and 94.6\% of
possibly solvable instances in experimental results presented in Table
\ref{tbl:sunny_comparison} and Table \ref{tbl:proteus_comparison}.
Considering the closeness to the oracle, we consider all of the
evaluated approaches to be state-of-the-art.

% ------------------------------------------------------------------------------
\subsection{The effect of the solving timeout on the portfolio
  effectiveness}
\label{sec:autocv}
%------------------------------------------------------------------------------
The bulk of the time for building the \argocspknn{} portfolio is spent
in the preparation phase (instance solving) and this time is directly
dependent on the value of the solving timeout used in that
phase. Since this is performed only once, usually we are prepared to
allow more time for preparation in order to achieve better results in
the portfolio exploitation. However, we wanted to examine the effect
of the solving timeout on the overall effectiveness of \argocspknn. In
all cases, the effectiveness is evaluated using 5-fold cross
validation over the whole corpus. The preparation is repeated several
times with different values of solving timeouts, ranging from 1 to 600
seconds, and to get a fair comparison, during cross-validation the
corpus was always partitioned to the same 5 folds. Regardless of the
solving timeout used in preparation, the time given to the selected
solver during exploitation was 600 seconds.

\begin{table}[!h]
  \begin{center}
    \resizebox{\textwidth}{!}{
      \begin{tabular}{|l|c|c|c|c|c|c|c|c|c|}\hline
        Solving timeout (seconds) & 1    & 5    & 10   & 30   & 60   & 90   & 120  & 300   & 600\\\hline
        Inst. solved in exploitation          & 6918 & 7051 & 7103 & 7328 & 7392 & 7422 & 7469 & 7495  & 7511\\\hline
        Prep. \& train. time (days) & 1.4  & 5.8  & 10.8 & 28.9 & 53.8 & 77.2 & 99.7 & 219.3 & 393.2 \\\hline
        Exploitation time (days)  & 11.8 & 11.0 & 10.5 & 8.8  & 8.4  & 8.2  & 8.0  & 8.0   & 7.9\\\hline
        Total time (days)          & 13.2 & 16.8 & 21.3 & 37.7 & 62.2 & 85.4 & 107.7& 227.2 & 401.2\\\hline
      \end{tabular}
    }
  \end{center}
  %\addtolength{\belowcaptionskip}{-8mm}
  \addtolength{\abovecaptionskip}{-2mm}
  \caption{Results of experimental evaluation of \argocspknn{} using
    different solving timeouts (given in seconds). Time spent is given
    in days.}
  \label{tbl:autocv}
\end{table}

\begin{figure}
  \begin{tikzpicture}[scale=0.8]
 ]
    \begin{axis}[
xlabel=Solving timeout per instance (seconds), ylabel=Number of solved instances,
]
      \pgfplotsset{
        legend style={at={(1,1)},anchor=north west,font=\small},
      }
      \addplot [color=black,mark=+,mark size=2pt] coordinates{
        (1,6918) (2,6961) (3,7007) (4,7035) (5,7051) (10,7103) (15,7145) (20,7197) (25,7228) (30,7328) (45,7375) (60,7392) (90,7422) (120,7469) (180,7472) (300,7495) (450,7500) (600,7511)
      };\addlegendentry{\argocspknn}
      \addplot [color=black,mark=--,mark size=2.5pt, dashed] coordinates{
        (0,6216) (600,6216)
      };\addlegendentry{\bestfixed{}}
      \addplot [color=black,mark=-,mark size=1.5pt] coordinates{
        (0,7631) (600,7631)
      };\addlegendentry{\oracle}
               {Case 1,Case 2}
    \end{axis}
  \end{tikzpicture}
  \caption{Results of experimental evaluation of \argocspknn{}
    compared to the \bestfixed{} solver and \oracle; each mark on
    curve represents one outcome based on solving using the
    corresponding timeout.}
  \label{fig:autocv}
\end{figure}
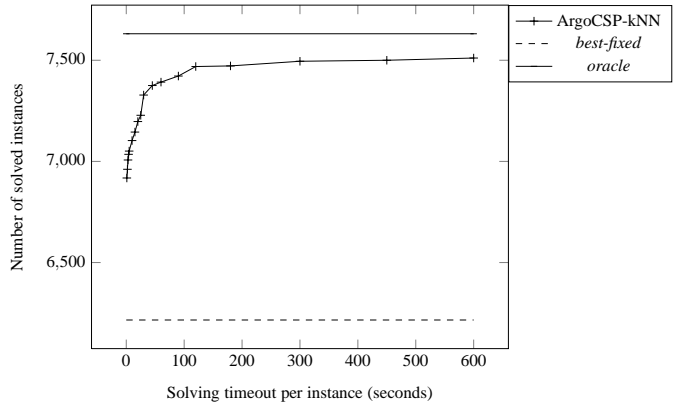

Table \ref{tbl:autocv} shows the obtained results and they are plotted
in Figure \ref{fig:autocv} (with results for additional values of
timeouts). The figure clearly shows that in the beginning there is a
sharp increase in the number of solved instances with the increase of
solving timeout, but after certain point the curve stabilizes. E.g.,
for solving timeouts of 600 and 30 seconds the difference in number of
solved instances is not very big (especially when their performance is
compared to method \bestfixed{}), while the time used for preparation
and training reduces from 393.2 to 28.9 days.
This demonstrates that it is possible to achieve almost the same
results even with much shorter solving timeouts.

% ------------------------------------------------------------------------------
\section{Short Portfolio Training}
\label{sec:short_training}
%------------------------------------------------------------------------------
Users often aim at solving as many instances from a single fixed
set of instances, as fast as possible, with the given set of available
solvers. A straightforward way would be to choose a single solver that
is expected to give the best results (e.g., choose the solver that
gave the best results on some solver competition) and apply it on all
instances. However, it is not clear how to choose a solver and the
chosen solver might not be suited to the specific set of instances.
Another approach would be to use a previously trained
portfolio. However, the portfolio might have been trained on the set
of instances of significantly different properties than the ones that
need to be solved, so it might perform poorly.

Encouraged by the results of the previous section that showed that one can
significantly reduce the solving timeout without significantly reducing
the overall portfolio quality, we propose a new way to efficiently use
portfolio approaches in situations when there is a need for solving a
previously unseen corpus as fast as possible, using a set of available
solvers. The approach relies on short portfolio training on the very same corpus
of instances that need to be solved. Due to the nature of the problem
(only solving single fixed corpus is important), the generalization of
the portfolio to instances out of this corpus is not required, so the
overlapping between the training and the test set of instances, which
is most common concern in machine learning, becomes irrelevant and
will be disregarded in the rest of this section.

The approach is simple and it consists of applying all solvers to all
instances in the corpus with a given short solving timeout,
performing predictive model training, and then applying the portfolio to
the corpus containing only instances not solved during
preparation, giving the selected solvers higher
solving timeout in exploitation phase.

\subsection{Evaluation}
\begin{figure}
  \begin{tikzpicture}[scale=0.8]
 ]
    \begin{axis}[
xlabel=Solving time per instance (seconds), ylabel=Number of solved instances,
]
      \pgfplotsset{
        legend style={at={(1,1)},anchor=north west,font=\small},
      }
      \addplot [color=black,mark=x,mark size=2pt] coordinates{
%(1,6919) (2,6971) (3,7019) (4,7056) (5,7065) (10,7117) (15,7158) (20,7222) (25,7256) (30,7356) (45,7410) (60,7427) (90,7463) (120,7511) (180,7522) (300,7562) (450,7574) (600,7578) % Old plot (with running selected solver again)
        (1,6925) (2,6987) (3,7044) (4,7069) (5,7088) (10,7145) (15,7306) (20,7356) (25,7377) (30,7377) (45,7424) (60,7444) (90,7490) (120,7536) (180,7571) (300,7604) (450,7618) (600,7631)
      };\addlegendentry{Short training}
      \addplot [color=black,mark=--,mark size=2.5pt, dashed] coordinates{
        (0,6216) (600,6216)
      };\addlegendentry{\bestfixed{}}
      \addplot [color=black,mark=-,mark size=1.5pt] coordinates{
        (0,7631) (600,7631)
      };\addlegendentry{\emph{oracle}}
               {Case 1,Case 2}
    \end{axis}
  \end{tikzpicture}
  \caption{Results of evaluation of \argocspknn{} using short training
    compared to the \bestfixed{} solver and \emph{oracle}. Each mark
    on the curve represents one outcome based on corresponding solving timeout.}
  \label{fig:autost}
\end{figure}
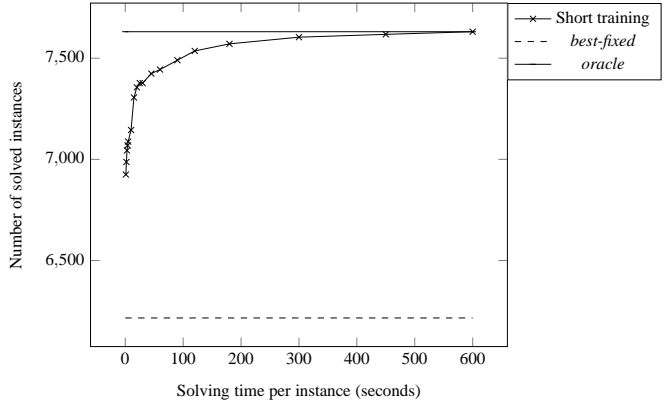

In the evaluation, we used different solving timeouts in training
phase, but due to the nature of our scenario (we cannot afford long
solving time), we focused on using very short timeout values. The
results are shown in Table \ref{tbl:autost} and plotted in Figure
\ref{fig:autost}. The total number of solved instances rapidly
increases in the region of short solving timeouts and then
stabilizes. For the solving timeout of 600 seconds, there is no need
to run selected solver on instances not solved in preparation phase
with that very same timeout (everything is done in the preparation
phase and the number of solved instances is the same as for the
\oracle{} solver).

\begin{table}
  \begin{center}
    \resizebox{\textwidth}{!}{
      \begin{tabular}{|l|c|c|c|c|c|c|c|c|c|c|}\hline
        Solving timeout (seconds)       & 1    & 5    & 10   & 30   & 60   & 90   & 120  & 300  & 600\\\hline
        Solved instances                & 6925 & 7088 & 7145 & 7377 & 7444 & 7490 & 7536 & 7604 & 7631\\\hline
        Prep. \& training time (days)      & 1.4  & 5.8  & 10.8 & 28.9 & 53.8 & 77.2 & 99.7 & 219.3& 393.2 \\\hline
        Exploitation time (days)  & 11.8 & 10.6 & 10.1 & 8.2  & 7.5  & 7.1  & 6.8  & 6.1  & 0\\\hline
        Total time (days)               & 13.2 & 16.4 & 20.9 & 37.1 & 61.3 & 84.3 & 106.5& 225.4& 393.2\\\hline
      \end{tabular}
    }
  \end{center}
  %\addtolength{\belowcaptionskip}{-8mm}
  \addtolength{\abovecaptionskip}{-2mm}
  \caption{Results of evaluation of \argocspknn{} using short training approach for different solving timeouts (given in seconds). The total time
    used is given in days.}
  \label{tbl:autost}
\end{table}

The results in tables \ref{tbl:autocv} and \ref{tbl:autost} and curves
in figures \ref{fig:autocv} and \ref{fig:autost} might seem
similar. The key difference is that the evaluation in Table
\ref{tbl:autocv} and Figure \ref{fig:autocv} is done by using
cross-validation and there is no overlapping between the training and
the testing set of instances.  Contrary to that, in the evaluation
shown in Table \ref{tbl:autost} and Figure \ref{fig:autost} the same
corpus is used both for training and testing (if an instance is solved
during training, it is not solved again during testing). Therefore,
the short training method constantly achieves better performance, but
this is expected as the training and the test sets consist of the same
set of instances.

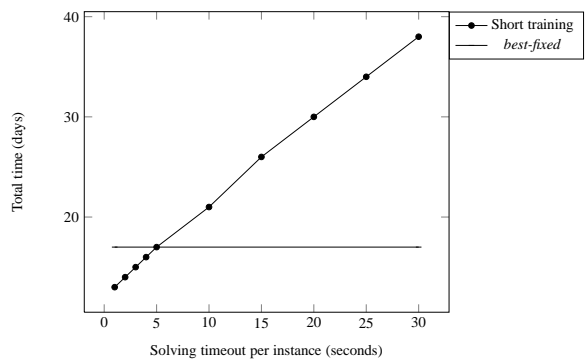
\begin{figure}[!h]
  \begin{tikzpicture}[scale=0.7]
 ]
    \begin{axis}[
xlabel=Solving timeout per instance (seconds), ylabel=Total time (days),
]
      \pgfplotsset{
        legend style={at={(1,1)},anchor=north west,font=\small},
      }
      \addplot [color=black,mark=*,mark size=1.5pt] coordinates{
           (1,13) (2,14) (3,15) (4,16) (5,17) (10,21) (15,26) (20,30) (25,34) (30,38)
      };\addlegendentry{Short training}
      \addplot [color=black,mark=-,mark size=1.5pt] coordinates{
        (1,17) (30,17)
      };\addlegendentry{\bestfixed{}}
               {Case 1,Case 2}
    \end{axis}
  \end{tikzpicture}
  \caption{Results of evaluation of \argocspknn using short training
    and the \bestfixed{} solver. Each mark on curve
    represents one outcome based on corresponding solving timeout.}
  \label{fig:autost_time}
\end{figure}

In the practical setting that we described, central quality measures
are the number of solved instances and the total time spent for both
training and exploitation -- one should try to maximize the number of
solved instances but keep the total time spent short (e.g., so that
both portfolio training and exploitation can be performed on a
computer that user has on his disposal). Surprisingly, our experiments
show that even for extremely small solving timeouts, the number of
solved instances is greater than for the \bestfixed{} solver (e.g.,
for the solving timeout of only 1 second, the number of solved
instances increases from 6216 to 6925, while the total time reduces
from 16.9 to 13.2 days, where 1.4 days are used for training and 11.8
are used for exploitation). In approximately the same time used by
the \bestfixed{} solver to solve 6216 instances, our approach solves 7088
instances (for solving timeout of 5 seconds). The relation between
solving timeout and total time used is presented in Figure
\ref{fig:autost_time}). The figure shows that total time used grows
linearly with the increase of the timeout. We argue that in practical
scenario it makes sense to use solving timeout of up to 30 seconds,
depending on the available time, because the number of solved
instances does not grow as fast for larger timeout values.

Another important question arises --- how much performance do we lose
using the proposed approach compared to the situation in which we already
have a portfolio built on the instances of similar properties to the
instances being solved? A good estimate is readily provided by comparison
with the results of evaluation performed in the previous section
and presented in Table \ref{tbl:autocv}.
If \argocspknn{} was trained with the solving timeout of 600 seconds,
it would need 393.2 days for training and it would solve 7511 instances.
On the other hand, if \argocspknn{} was used with the short training method
with the solving timeout of 60 seconds and then with exploitation
timeout of 600 seconds per instance, the portfolio would need
53.8 days for training and it would solve 7444 instances (Table
\ref{tbl:autost}). The difference of 11 months in training time is
obviously very significant and the difference in number of solved instances, maybe
not as much, but that is left to the user to decide, based on the purpose
and the time available.

\subsection{Comparison of different machine learning techniques}

K-nearest neighbors is one of the simplest machine learning
techniques.  Therefore, it is interesting to perform a comparison to
some other machine learning techniques and see if k-nearest neighbors
should be substituted by a more promising method in the context of
short-training. The way we use k-nearest neighbors algorithm in
\argocspknn{} is a bit specific to particular application, and not all
machine learning algorithms can be applied in the same way. Therefore,
in the comparison we use a bit different, but perfectly natural
portfolio design (e.g. SATzilla uses such design \citep{satzilla}), and
also include \argocspknn{} as it is presented in previous subsection.

The portfolio design used in this subsection is based on prediction of
solving time. In the exploitation phase, for each solver, its runtime
on input instance is predicted by previously learned model and the
solver with the least predicted runtime is invoked to solve the input
instance. In the model training phase, the features of all training
instances and PAR10 scores of all CSP solvers on all of these
instances are used to train the model for each solver. In order to
build a model, several combinations of values of metaparameters are
evaluated on the training set using 5-fold cross-validation. The best
combination is used to build a model on the training set for each
solver.

We conducted experiments with regression machine learning techniques
to evaluate their efficiency. We experimented with: \knn, linear
regression and SVM. Tool
\rapidminer{}\footnote{\url{https://rapidminer.com/}} was used for the
experiments. As \rapidminer{} performed very poorly when used with SVM
(the process just got stuck with many different combinations of
metaparameters), we also experimented with tool \libsvm{}
\citep{libsvm} which showed greater stability. We used two types of
kernels with SVM: linear and RBF.

For the efficiency reasons, we used subset of 28 features from 70
features (we eliminated almost half of the features because they had
value 0 for all instances and a few features that were in our opinion
less important). We used 5-fold cross-validation. The training set and
the testing set were normalized before usage. Given a training set,
all its instances are solved using each of the included solvers in a
given solving timeout, and then the optimal metaparameters are
selected in the way already described. The metaparameters for \knn{}
were the number of neighbors $k$ (ranging from 1 to 20) and the
distance measure $d$ (4 different distance measures described by
Tomovi\' c et al. \citep{distance_functions}). The metaparameters for
SVM using linear kernel were \emph{C} (powers of 2 from $2^{-6}$ to
$2^{15}$), \emph{$\nu$} (ranging from $0.1$ to $0.9$ with step
$0.1$). For RBF kernel we used the same combinations of metaparameters
as for linear kernel and additionally metaparameter $\gamma$ (powers
of 2 from $2^{-15}$ to $2^5$). The metaparameters for linear
regression were \emph{ridge} (powers of 10 from $10^{-8}$ to
$10^{-1}$), \emph{use\_bias} (true or false),
\emph{feature\_selection} (none, M5 prime or greedy) and
\emph{eliminate\_colinear\_features} (true or false). For each method,
all combinations of metaparameters are tried on the training set and
the ones generating the best score are declared optimal.

Table \ref{tbl:ml_techniques} shows the obtained results. The \knn{}
method solves significantly more instances than other used approaches,
and \argocspknn{} solves even more (its performance with the reduced
set of features is almost the same as with the full set, already
presented in Table \ref{tbl:autost}). As idea is to compare different
machine learning techniques using short training, we had to limit the
time used for prediction. For both kernels in libsvm tests, months
would be needed to finish these experiments, so we used the reduced
sets of metaparameter values giving the best results in preliminary
experiments. Even with these reduced set of parameters libsvm tests
with RBF kernel needed more than a month to finish. We therefore
conclude that due to time consumption, SVM models are not suitable for
short training approach.

\begin{table}[!h]
  \begin{center}
    {
      \begin{tabular}{|l|c|c|c|c|c|c|c|c|c|}\hline
        & 1    & 5    & 10   & 30   & 60\\\hline
        Linear regression (RM) & 5947 & 5950 & 6073 & 6291 & 6459 \\\hline
        Linear kernel (libsvm) & 6216 & 6753 & 6753 & 6762 & 6695 \\\hline
        RBF kernel (libsvm)    & 6235 & 6663 & 6518 & 6716 & 6748 \\\hline
        \knn{} (RM)            & 6052 & 6644 & 6804 & 7201 & 7309 \\\hline
        \knn{} (\argocspknn{}) & 6927 & 7095 & 7147 & 7375 & 7458 \\\hline
%        \knn{} (\argocspknn{}) & 6897 & 7058 & 7100 & 7317 & 7404 \\\hline % Ovo je CV approach
      \end{tabular}
    }
  \end{center}
  %\addtolength{\belowcaptionskip}{-8mm}
  \addtolength{\abovecaptionskip}{-2mm}
  \caption{Results of experimental evaluation using different machine
    learning techniques and different solving timeouts.}
  \label{tbl:ml_techniques}
\end{table}

As in subsection \ref{sec:autocv}, we estimate performance of
portfolio previously trained on similar (but not the same) set of
instances, using 5-fold cross-validation on the corpus with different
machine learning techniques and solving timeout of 600
seconds. Results are given in Table \ref{tbl:ml_techniques_600}. In this
case, no reduction of metaparameters in case of SVM was performed, but
still, both versions of \knn{} perform better than other approaches.

\begin{table}[!h]
  \begin{center}
    {
      \begin{tabular}{|l|c|}\hline
        & Solved instances \\\hline
        \bestfixed{}           & 6216 \\\hline
        Linear regression (RM) & 6515 \\\hline
        Linear kernel (libsvm) & 6715 \\\hline
        Rbf kernel (libsvm)    & 7323 \\\hline
        \knn{} (RM)            & 7432 \\\hline
        \knn{} (\argocspknn{}) & 7511 \\\hline
        \oracle{}              & 7631 \\\hline
      \end{tabular}
    }
  \end{center}
  %\addtolength{\belowcaptionskip}{-8mm}
  \addtolength{\abovecaptionskip}{-2mm}
  \caption{Results of evaluation using 5-fold cross-validation and different machine learning techniques.}
  \label{tbl:ml_techniques_600}
\end{table}

% ------------------------------------------------------------------------------
\subsection{Evaluation on SAT Instances}
\label{sec:sat}
%------------------------------------------------------------------------------

Since our short training methodology gave very good results with CSP
portfolio inspired by \argosmart{}, it is interesting to show that our
short training methodology is applicable to SAT corpora, as
well. Therefore we run experiments on SAT instances using original
\argosmart{} portfolio.

\begin{figure}
  \begin{tikzpicture}[scale=1]
 ]
    \begin{axis}[
xlabel=Solving time per instance (seconds), ylabel=Number of solved test instances,
]
      \pgfplotsset{
        legend style={at={(1,1)},anchor=north west,font=\small},
      }
      \addplot [color=black,mark=x,mark size=2pt] coordinates{
%        (1,3275) (2,3326) (3,3329) (4,3388) (5,3413) (10,3495) (15,3569) (20,3597) (25,3633) (30,3645) (45,3701) (60,3730) (90,3777) (120,3790) (180,3826) (300,3892) (450,3896) (600,3924) % Old plot
        (1,3286) (2,3342) (3,3347) (4,3399) (5,3426) (10,3521) (15,3599) (20,3632) (25,3646) (30,3679) (45,3731) (60,3763) (90,3815) (120,3826) (180,3877) (300,3927) (450,3978) (600,4000)
      };\addlegendentry{Short training}
      \addplot [color=black,mark=-,thick,mark size=1pt, dotted] coordinates{
        (0,3822) (600,3822)
      };\addlegendentry{Cross-validation}
      \addplot [color=black,mark=--,mark size=2.5pt, dashed] coordinates{
        (0,2490) (600,2490)
      };\addlegendentry{\bestfixed{}}
      \addplot [color=black,mark=-,mark size=1.5pt] coordinates{
        (0,4000) (600,4000)
      };\addlegendentry{\oracle}
               {Case 1,Case 2}
    \end{axis}
  \end{tikzpicture}
  \caption{Results of evaluation of \argosmart{} portfolio on SAT instances
    using short and standard training, \bestfixed{}
    solver and \oracle. Each mark on curve represents an
    outcome based on corresponding solving
    timeout.}
  \label{fig:autosat}
\end{figure}
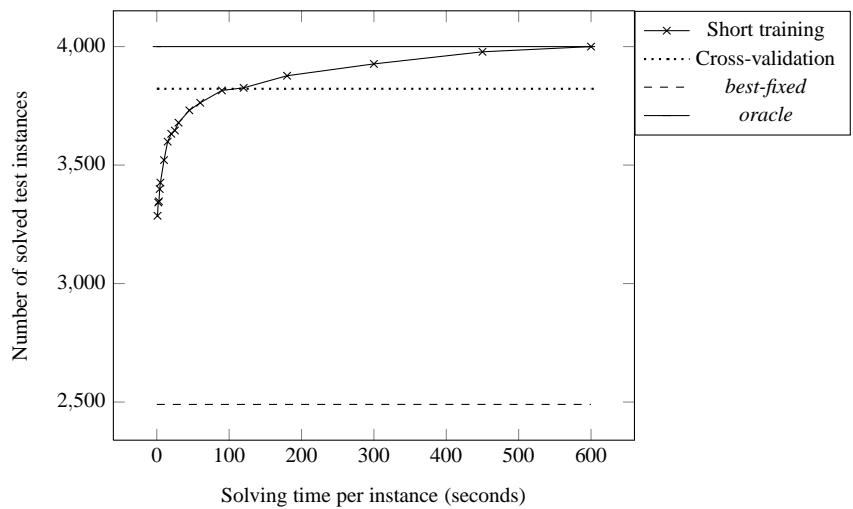

We compare \argosmart{} with and without short training methodology on
instances originally used in \argosmart{} paper \citep{nikolic2}
-- instances from SAT Competitions (2002-2007) and SAT Races (2000-2008).
In short training approach, portfolio is both trained and
run on those instances. In original approach without short training,
the portfolio is evaluated using 5-fold cross-validation. The results
of the comparison are given in Figure \ref{fig:autosat}. Since in the
original approach there is no sequence of timeouts, it is represented
as a line parallel to horizontal axis. The curve for short training
approach looks very similar to the one presented in previous
subsection. Very good results are obtained for very small
timeouts. Therefore, our methodology can be efficiently applied not
only to CSP instances, but also to SAT instances.

% ------------------------------------------------------------------------------
\section{Conclusions and Further Work}
% ------------------------------------------------------------------------------
\label{sec:conclusions}
%------------------------------------------------------------------------------
In this paper, we have presented a simple \knn{}-based portfolio for
CSP and shown that it achieves state-of-the-art performance by
comparison with other methods.  We have assessed the effects of the
training phase duration to overall portfolio quality by using
different solving timeout values. Our approach significantly improves
over each single constituent solver and gives good generalization
results even when used with a very short preparation phase (so that
the training does not require an advanced cluster computer, but can be
done on a single PC).

Often, solving a single fixed set of instances as fast as possible
with the given set of solvers is the only practical concern. We have
addressed this problem and proposed a portfolio-based approach that
uses short portfolio training on the set of instances to be solved and
then selects a suitable solver for each of those instance based on the
learned predictive model. Experimental results indicate that this
approach significantly improves over the best fixed solver, even for
very short timeouts. Also, the number of solved instances first
quickly increases when increasing the solving timeout value, but then
saturates, and from some point, increasing the solving timeout yields
small number of additional solved instances. Therefore, one can choose
a solving timeout depending on the time he has available and still
significantly improve the results over any single solver.

We also performed experiments with SAT instances and our methodology
also shows good results on this kind of problems.

A possible direction in improving short training approach could be to
automatically select the solving timeout for the set of instances to
be solved, based on the total time available. This would make the
approach more autonomous and the one would not need to think about the
solving timeout that is going to be used, but to simply determine the
total time available. Also, it would be good to consider configuration
options of a diverse set of solvers and while choosing a solver also
to choose its suitable configuration. For example, systems \mesat{},
\azucar{}, and \bee{} \citep{bee} offer choosing between different
encodings, while Minisat++ \citep{minisatplusplus} offers 3 different
options when encoding PB constraints (adders, BDDs, Sorters). One more
direction of future work, would be to test our approach on other types
of problems (e.g. Answer Set Programming, Constraint Optimization
Problems).

\begin{acknowledgements}
This work was partially supported by the Serbian Ministry of Science
grant 174021.
\end{acknowledgements}

%------------------------------------------------------------------------------
\label{sect:bib}
\bibliographystyle{spbasic}      % basic style, author-year citations
\bibliography{portfolio}

%------------------------------------------------------------------------------
\end{document}